\documentclass[]{interact}

\usepackage{epstopdf}
\usepackage[caption=false]{subfig}

\usepackage[natbibapa,nodoi]{apacite}
\renewcommand\bibliographytypesize{\fontsize{10}{12}\selectfont}

\theoremstyle{plain}

\theoremstyle{definition}

\theoremstyle{remark}

\usepackage{array}
\usepackage{multirow}

\begin{document}


\title{GeoAI Reproducibility and Replicability: a computational and spatial perspective}

\author{
\name{Wenwen Li\textsuperscript{a,*}\thanks{CONTACT Wenwen Li. Email: wenwen@asu.edu}, Chia-Yu Hsu\textsuperscript{a}, Sizhe Wang\textsuperscript{a,b}, Peter Kedron\textsuperscript{c}}
\affil{\textsuperscript{a}School of Geographical Sciences and Urban Planning, Arizona State University, Tempe, AZ, USA, 86287-5302; \textsuperscript{b}School of Computing and Augmented Intelligence, Arizona State University, Tempe, AZ, USA, 86287-5302; \textsuperscript{c}Department of Geography, University of California, Santa Barbara, 93106-4060}
}

\maketitle

\begin{abstract}
GeoAI has emerged as an exciting interdisciplinary research area that combines spatial theories and data with cutting-edge AI models to address geospatial problems in a novel, data-driven manner. While GeoAI research has flourished in the GIScience literature, its reproducibility and replicability (R\&R), fundamental principles that determine the reusability, reliability, and scientific rigor of research findings, have rarely been discussed. This paper aims to provide an in-depth analysis of this topic from both computational and spatial perspectives. We first categorize the major goals for reproducing GeoAI research, namely, validation (repeatability), learning and adapting the method for solving a similar or new problem (reproducibility), and examining the generalizability of the research findings (replicability). Each of these goals requires different levels of understanding of GeoAI, as well as different methods to ensure its success. We then discuss the factors that may cause the lack of R\&R in GeoAI research, with an emphasis on (1) the selection and use of training data; (2) the uncertainty that resides in the GeoAI model design, training, deployment, and inference processes; and more importantly (3) the inherent spatial heterogeneity of geospatial data and processes. We use a deep learning-based image analysis task as an example to demonstrate the results' uncertainty and spatial variance caused by different factors. The findings reiterate the importance of knowledge sharing, as well as the generation of a ``replicability map'' that incorporates spatial autocorrelation and spatial heterogeneity into consideration in quantifying the spatial replicability of GeoAI research.
\end{abstract}


\section{Introduction}
Geospatial artificial intelligence, or GeoAI, has become a trending research topic in GIScience. GeoAI is defined as the transdisciplinary expansion of AI into the geospatial world \citep{liGeoAIWhereMachine2020}, and is generally comprised of a set of machine learning and deep learning techniques designed to extract useful information from structured and unstructured geospatial data. Current enthusiasm about AI techniques in the field of geography is, in part, due to the increasingly data-intensive nature of the discipline and the challenges that intensity brings to conventional forms of spatial analysis. As Earth observation satellites, distributed environmental sensor networks, and global positioning systems continue to create and expand already massive location-based datasets at unprecedented speeds, researchers face the challenge of leveraging this data when addressing pressing social and environmental problems \citep{vopham2018emerging, li2022geoai,liu2022review, usery2022geoai}, . Emerging deep learning techniques have already demonstrated a promising capacity to extract information from big geospatial data by processing and analyzing geospatial data at previously unprecedented spatial and temporal scales with high accuracy. As the volume, velocity, and variety of geospatial data continue to increase, there remains an urgent need to continue to develop and improve GeoAI methodologies so they can be brought to bear on pressing problems \citep{gao2023handbook, li2023assessment}. 

As GeoAI has emerged, a growing number of studies have incorporated these techniques, especially deep learning methods, into their research design and analyses.
The expanding use of GeoAI has raised questions about how to best facilitate the reproducibility and replicability of this work \citep{goodchildReplicationSpaceTime2021,kedronReproducibilityReplicabilityOpportunities2021,  wilsonFivestarGuideAchieving2021}. 
Attempting to reproduce or replicate research plays and important epistemological role in scientific inquiry. 
Broadly, reproductions and replications provide diagnostic evidence about the validity of the claims made in prior studies, or seek to test and extend those claims to new contexts. 

\begin{figure}[hbt!]
\centering
\includegraphics[]{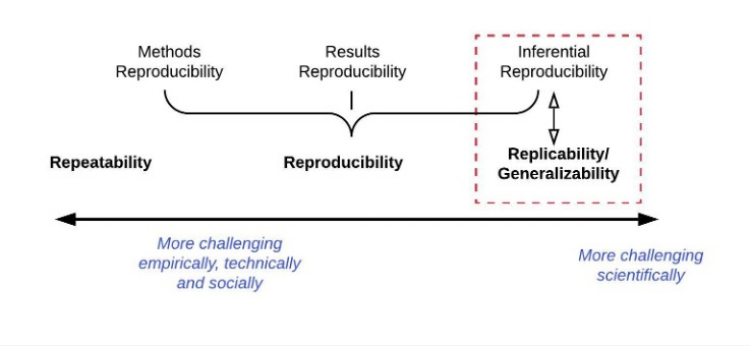}
\caption{The R\&R spectrum in GeoAI research.} \label{fig_rrspectrum}
\end{figure}

Following the \cite{associationforcomputingmachineryArtifactReviewBadging} at least three forms of reproducibility are relevant to computationally intensive analyses such as GeoAI (Figure \ref{fig_rrspectrum}).  First, ``repeatability'' emphasizes whether the same results can be derived by the same researcher using the same data and the same experimental procedure across multiple trials. In the context of spatial data analysis and GeoAI, the ``same experimental procedure'' means the use of the same software or library on the same computer with exactly the same parameter settings. Studies designed to test the repeatability of prior research offer the simplest form of reproducibility check by providing evidence that the results of a computation are not the product of some internal error or property of the computational setup. 

Second, the ``reproducibility'' of the results of a computational analysis refers to the capability of a independent researcher to obtain the same results using the same data as an earlier study and similar procedures. 
Reproducibility can be further categorized into methods reproducibility and results reproducibility by distinguishing the focus of the the reproduction\citep{goodmanWhatDoesResearch2016}. Methods reproducibility is similar to repeatability, in that it centers on whether enough information and detail were provided about the data and methods of the original study to allow the same procedures to be repeated. Methods reproductions try to verify research finding using the same data and same analysis (e.g., same method in the same software), but are conducted by a different researcher perhaps in a different experimental environment (e.g., different computer). In the context of GeoAI, a methods reproduction would control for the influence of the original researcher and the computational environment. In contrast, results reproducibility focuses on whether the results remain consistent when a second researcher closely matches, but adjusts, the methods used in the initial study and uses the same data (same/similar method, different software). Compared to repeatability which simply verifies whether research is repeatable and yields the same outcome, results and methods reproducibility provide evidence that can be used to assess the reliability of results. Reproducibility, in all of its forms, is an essential step in the development of GeoAI techniques because GeoAI research typically adapts and extends state-of-the-art (SOTA) models from general AI into the geospatial domain. As such, verifying the results using the original model and data reported in a foundational AI paper, the extension of that model to the geospatial domain, and on through its further adaptations is critical to facilitating the creation of credible and reliable methodologies.

Third, ``replicability'', and the analogous concept of inferential reproducibility, refers to the capability of an independent researcher to draw same conclusion using similar methods to analyze a different set of data. Replication provides evidence that can be used to assess the external validity of the claims of a study. In the context of geographic research and GeoAI, conducting replications is a means of assessing whether the conclusions drawn using data from one geographical area are supported when tested with data from another location. This emphasis on evaluating the impact of changes in location on research findings distinguished replicability from reproducibility and repeatability.

While reproducibility and replicability have become increasingly prominent in geography, research efforts have primarily focused on repeatability and computational forms of methods or results reproducibility \citep{konkol2019}. In contrast, relatively few studies have investigated the replicability of research findings across space. This absence is somewhat surprising as replication of findings across space is arguable a central question of the discipline \citep{kedron2022replication,sui2021reproducibility}. \citet{goodchildReplicationSpaceTime2021} argued that these studies may be missing because research in social and environmental science can only be weakly replicable due to the spatial heterogeneity that ubiquitously resides in spatial data and spatial processes. That issue may be further amplified in GeoAI research because many of these techniques remain difficult to interpret and ``black box'' even without the additional complication of a change in location \citep{hsu2023explainable}. To work toward a solution, Li and Goodchild call for research that might generate a ``replicability map'' that better quantifies the impact of location on the R\&R of GeoAI models. This paper responses to this call by using crater detection over the global Mars surface as a case study to investigate the reproducibility and replicability of GeoAI from a computational and spatial perspective. 

The remainder of the paper is organized as follows: Section 2 summarizes challenges that currently hinder R\&R in GeoAI research. Section 3 introduces the methodology, including data, model, and experimental design, to measure the computational and spatial variances in a GeoAI model’s predictions. Section 4 presents the results and provides a detailed analysis of how the computational settings of the model (training data size and random seeds) as well as location variance (e.g., spatial autocorrelation and spatial heterogeneity) affect the R\&R in GeoAI. Section 5 concludes the paper and proposes future research directions.

\section{Computational and spatial challenges toward achieving R\&R in GeoAI research}
\subsection{Computational challenges in achieving R\&R in GeoAI }
GeoAI models, especially deep learning models, are known for the complexity of their architecture design, training methods, and hyperparameters setting \citep{maxwellEnhancingReproducibilityReplicability2022}.  When compared to traditional statistical methods or shallow machine learning models, this complexity expands the number of decisions a researcher developing or using deep learning methods must make when conducting a study, which makes ensuring the repeatability and reproducibility of a work more challenging to achieve. Table \ref{tbl_rrchallenges} summarizes these factors and additional challenges that arise when using heterogeneous computing environments with frequent software updates. Besides model complexity, deep learning models also often employ random seeds \citep{phamProblemsOpportunitiesTraining2020} in their learning process. For example randoms seeds are used when assigning initial model weights, conducting data shuffling to avoid overfitting, and to search for globally optimal solutions. When seeds are not set to fixed value an additional variation is introduced into the modelling process and repeating or reproducing the origina results become more difficult.

\begin{table}[hbt!]
\tbl{Computational and modeling challenges that affect GeoAI R\&R.}
{\begin{tabular}{p{0.18\textwidth} p{0.45\textwidth} >{\centering\arraybackslash}p{0.12\textwidth} >{\centering\arraybackslash}p{0.13\textwidth}} 
\toprule
Computational/ modeling challenges & Description & Affecting repeatability? & Affecting reproducibility?  \\ 
\midrule
Intricate GeoAI model architecture & Vague descriptions in the construction of the model, such as insufficient information about the model backbone, including the number of convolutional or attention blocks and associated components (e.g., use of batch normalization and activation function), as well as the utilization of the loss function. & N/A & Yes \\
Model uncertainty &	Failure to fix random seeds for model weights initiation or data shuffling.  & Yes & Yes \\
Inference methods &	Uncertainty and randomness introduced in inference methods, such as beam search and temperature-scaled sampling, adopted in the heuristic-based data processing and inference stages. & Yes & Yes \\

Use of training data & Vague or lack of description in input data dimension/ input sample size, data split among training, validation, and testing dataset. Issues of dataset availability, especially when datasets are proprietary, too large, or dynamic. & N/A & Yes \\
Training methods & The unavailability of training code, which includes settings for and the process of tuning hyperparameters such as batch size, learning rate, the use of early stopping strategies, optimization algorithms, training iterations (epochs), and the application of data augmentation techniques, poses a challenge for reproducibility. & Yes & Yes \\
 Software/library & Constant auto-updates in machine learning libraries and their embedded functions & Yes & Yes \\ 
Computing environment & Changes in GPU architecture, floating-point discrepancies, and non-deterministic behaviors in GPUs. & N/A & Yes \\

\bottomrule
\end{tabular}}
\label{tbl_rrchallenges}
\end{table}

The uncertainty involved in the inference stage of GeoAI models may also increase the challenges for achieving R\&R. In natural language processing (NLP) research in particular, there are strategies developed toward the search for global optimal solutions by adding uncertainty in the heuristic-based data processing and inferences. For example, beam search \citep{meister2020if} is a strategy that improves upon greedy search by keeping track of multiple promising solutions at each step, rather than making the single best choice. At each step, it considers the top-k best options based on the model's current performance. Then, the next steps will involve trying out each of these options, allowing the model to explore a much larger search space toward an optimized solution. Similarly, temperature-scaled sampling \citep{shih2023long} is a strategy that fine-tunes the balance between predictability and creativity in text generation. By adjusting the ``temperature'', it influences how the model chooses the next word in a sequence. At a low temperature, the model favors more common and expected word choices. On the other hand, a high temperature setting encourages the model to select more diverse words and introduces a higher degree of creativity. Simulated annealing is a similar technique that is widely adopted in heuristic-based spatial problem-solving, such as regionalization \citep{li2014p}, beyond NLP. Other dynamic search strategies, such as top-k sampling \citep{kool2019stochastic} and Nucleus sampling \citep{Holtzman2020The} adopt different sampling strategies to determine the next step in the model's actions. By increasing the number and variations in its choices, the model may possibly be capable of rendering a better result than those with deterministic behaviors. However, this modeling approach also raises the uncertainty of the results and therefore challenges in reproducing the exact results from previous model runs.

Many of the computational challenges inhibiting the reproducibility of GeoAI can be addressed, or largely alleviated, by adopting open science practices designed to increase the transparency of geospatial research \citep{robinsonCommonBarTropopause2014,wrightDigitalResilience2016}. Recent advances in the open science movement \citep{boultonOpenYourMinds2012,reyOpenGeospatialAnalytics2015} have expanded data and code sharing \citep{jasnyFosteringReproducibilityIndustryacademia2017}, the advancement and adoption of versioning software \citep{shaoWhenSpatialAnalytics2020}, and more stringent requirements of repeatability and reproducibility by science journals \citep{borgmanConundrumSharingResearch2012}. These changes have created an environment where it is inceasingly possible to create detailed documentation that describes, in technical language, the entire research process -- from the data used to the construction, training, fine-tuning, and deployment of a GeoAI model \citep{anselinMetadataProvenanceSpatial2014,wilsonFivestarGuideAchieving2021, kedron2021GA}. 

\subsection{Spatial challenges in achieving R\&R in GeoAI}
While advances and adoption of open science practices is beginning to take root in geography \citep{reyOpenGeospatialAnalytics2015}, systematic research programs that test the replicability and generalizability of research findings across space and time have yet to emerge in the discipline, or among studies developing and deploying GeoAI. The absence of such a program, and the difficulty in developing and deploying it, echoes the long-standing nomothetic-ideographic débat/debate in geography \citep{goodchildReplicationSpaceTime2021,barnes2022great, sui2021reproducibility}. On one hand, the Earth's surface exhibits uncontrolled variance and non-stationarity \citep{anselinWhatSpecialSpatial1989}, motivating researchers to develop ideographic approaches that focus on studying the specific constructs of a region and its social system instead of seeking general laws. This suggests that geospatial-focused research cannot be replicated when the study area changes. On the other hand, the establishment of Scientific Geography as a law-seeking science has fostered advances in quantitative spatial analytical methods with generalizable model constructs applicable across different geographical areas. Concurrently, methods like Geographically Weighted Regression \citep[GWR;][]{fotheringhamGeographicallyWeightedRegression2003} and multi-scale GWR \citep{fotheringhamMultiscaleGeographicallyWeighted2017} allow parameters in the models to change from location to location, striking a balance between seeking general laws and accounting for spatial variance. However, these methods are recognized for having only ``weak replicability'' \citep{goodchildReplicationSpaceTime2021}. 

The challenge of replication becomes even more acute in GeoAI research. GWR produces a two-dimensional grided parameter space, where values represents the positive or negative impact of independent predictors (such as unemployment rate) on a response variable (e.g., crime rate) in the geographical area. This parameter surface can be interpreted as a statistical indicator of the stability and replicability of a relationship across space at a given geographical scale. In comparison, GeoAI methods use complex model architectures, and employ learning processes that test a multitude of variable relations, that frequently create interpretation problems. Moreover, GeoAI techniques are often employed to address different problems than those addressed by a regression model like GWR. First, many GeoAI models are developed to handle big geospatial data, making it challenging to collect training data that covers the entire globe. Second, many research topics utilizing GeoAI focus on solving specific problems within a particular study area. Very few studies have examined the replicability of their models across different geographical regions.

To fill this gap, this paper conducts a series of experiments to understand how the reproducibility and replicability of GeoAI models are influenced by changing geographical locations. We experimented with several key computational issues, such as training data size and model uncertainty, as well as spatial factors, including latitude, longitude, and geographical proximity. The next section describes our methods, data, and experimental design.

\section{Data and Method}
\subsection{Dataset}
We use crater detection over the global Mars surface to assess how the reproducibility of GeoAI models if affected by changes in location. The Mars Crater Dataset \citep{hsuKnowledgeDrivenGeoAIIntegrating2021} is a comprehensive resource for spatial analysis and GeoAI applications. This dataset comprises 102,675 image scenes derived from the Mars global mosaic, which is based on the Mars Odyssey Thermal Emission Imaging System (THEMIS) daytime infrared data at 100m spatial resolution \citep{edwardsMosaickingGlobalPlanetary2011}. The dataset’s core strength lies in its detailed annotations: each image features instance-level bounding boxes marking craters, as identified in the extensive Mars Impact Crater Catalog by Robbins and Hynek \citep{robbinsNewGlobalDatabase2012}. This catalog, complied from extensive manual searches across infrared imagery and topographic data, documents over 640,000 craters, providing essential morphologic and morphometric data.

The careful selection of the 102,675 images from a potential total of 220,956, as conducted by \citet{hsuKnowledgeDrivenGeoAIIntegrating2021}, ensures the dataset's reliability for replicable scientific research. Each image scene covers an area of 25.6 km $\times$ 25.6 km at a size of 256 pixels by 256 pixels. This subset, selected and curated for its visual clarity and the recognizability of craters, contains a diverse range of crater sizes, ranging from as small as 0.2 km to as large as 25.5 km in diameter. This diversity ensures a comprehensive representation of the Martian landscape, enhancing the dataset's utility for consistent and replicable analyses. Originally developed for training and evaluating the effectiveness of GeoAI models for Mars crater detection, this dataset now serves as a foundational resource for our studies on the reproducibility and replicability of GeoAI.

\subsection{Tasks and model}
In our research, we delve into the domain of object detection, specifically tailored to the task of identifying craters on Mars. Object detection, a cornerstone of computer vision, aims to localize and categorize various objects within images \citep{li2021tobler}. This approach is particularly pertinent for our study, given the need for precise localization and characterization of craters across diverse Martian terrains. The complexity of this task stems from the varying sizes, shapes, and appearances of craters, coupled with the distinct geological features on the Mars surface. To tackle this problem, we employed a cutting-edge deep learning-based object detection model---MViTv2, which utilizes a vision transformer-based feature extraction backbone and multi-scale representation learning to achieve state-of-the-art performance in various image analysis and object detection tasks \citep{liMViTv2ImprovedMultiscale2022}. 

The MViTv2, an advanced iteration of the vision transformers \citep{dosovitskiyImageWorth16x162020,liExploringPlainVision2022} family, is distinguished by its enhanced capability to process and integrate information across multiple scales. This multiscale approach is critical for our task as it allows for the effective detection of craters of varying sizes, a fundamental requirement for accurate spatial analysis on Mars. Additionally, MViTv2's refined attention mechanism is especially effective in discerning subtle crater features, facilitating a more detailed understanding of the Martian surface. This attribute aligns closely with our research's emphasis on reproducibility and replicability, as it ensures consistent and reliable detection across different instances of the dataset. Moreover, the efficiency of MViTv2 in balancing computational demands and achieving high predictive performance makes it a suitable choice for extensive datasets like ours. This balance is essential for replicable and scalable GeoAI applications, ensuring that the model's performance is not hindered by the vastness or complexity of the data.

\subsection{Experimental design}
Our experimental design is structured around five distinct experiments, tailored to assess the reproducibility and replicability of GeoAI along both computational and spatial dimensions (Table \ref{tab:experiments}). The first two experiments focus on computational aspects: Experiment 1 examines the influence of training sample size on the variance of model performance and its reproducibility, while also seeking to establish the threshold of data volume necessary for obtaining robust results within the 102,675-image Mars Crater Dataset. Experiment 2 delves into the impact of fixing or varying random seeds on deep learning outcomes. By altering how random seed are managed during model implementation, we aim to understand the role of seed setting in ensuring deterministic behavior in training processes and the consequent stability of results---a key factor in ensuring computational reproducibility in GeoAI models.

The next three experiments explore the geographical attributes of the dataset and how GeoAI models’ generalizability is impacted by changing locations. Experiment 3 involves creating a grid-based ``replicability map'' by training a GeoAI model on randomly sampled data from the entire Martian surface and then applying this model across various grid cells. This approach is designed to reveal spatial variations in model performance, highlighting the importance of geographical diversity in the distribution of training data for replicable GeoAI results. Experiments 4 and 5 extend this spatial analysis further, examining latitude and longitude-induced variations in model performance, respectively. By training models with data from specific latitudes or longitudes and testing them on other parts of the Martian surface, we aim to uncover how prominent crater characteristics, such as density and shape, vary with geographical location and how these variations influence model performance across different regions of Mars. Collectively, these experiments provide a unique view of the challenges and considerations in achieving spatial replicability in GeoAI.

\begin{table}[hbt!]
\tbl{Computational and spatial emphasis of the R\&R experiments.}
{\begin{tabular}{llcccc} 
\toprule
 & Focus & Reproducibility & Replicability & Computational & Spatial \\ 

\midrule
Sec. 4.1 & Impact of training sample size & X & & X & \\
Sec. 4.2 & Random seed effect & X & & X & \\
Sec. 4.3 & Random location transfer & & X & X & X \\
Sec. 4.4 & Locational variance across varying latitude & & X & & X \\
Sec. 4.5 & Locational variance across varying longitude & & X & & X \\ 

\bottomrule
\end{tabular}}
\label{tab:experiments}
\end{table}

\section{Results Analysis}
\subsection{Impact of sample size on model stability and performance}
In the realm of modern deep learning, the quantity of training data is a crucial factor influencing model performance. The first experiment in our study aims to explore the impact of training sample size on the stability and effectiveness of our model. Our objective is to discern the relationship between data volume and model predictive accuracy, a critical aspect in ensuring computational reproducibility and reliability in GeoAI applications.

To conduct this experiment, we utilized varying subsets of the Mars Crater Dataset, specifically focusing on training sets of different sizes: 100, 200, 400, 800, 1,200, 1,600, 2,000, 2,400, and 2,800 images, respectively. The selection of these sizes is informed by our preliminary experimental testing. This structured approach allows us to observe how the model adapts and performs with varying amounts of data, helping us identify the optimal balance between data volume, model accuracy, computational efficiency, as well as the degree of results reproducibility as the size of training data changes. 

A consistent validation set of 10,000 images, representing approximately 10\% of the total dataset, was used across all training sizes for model evaluation and tuning. This choice of 10,000 images is designed to adequately capture the dataset’s diversity, ensuring statistically significant validation results while remaining computationally efficient for the model’s training process. It also allows for a uniform standard in evaluating the model’s performance, enabling direct comparisons between different training set sizes. Additionally, a testing set comprising another 10,000 images was used, completely separate from the training and validation datasets. This testing set, also representing about 10\% of the total dataset, is critical for an extensive and reliable evaluation of the model's ability to generalize to new, unseen data. The consistency in the size of both the validation and testing sets across various training scenarios is important, as it ensures that any variations in the model's performance can be clearly attributed to the differences in training sample size, rather than inconsistencies in evaluation criteria.

\begin{figure}
\centering
\includegraphics[width=\textwidth]{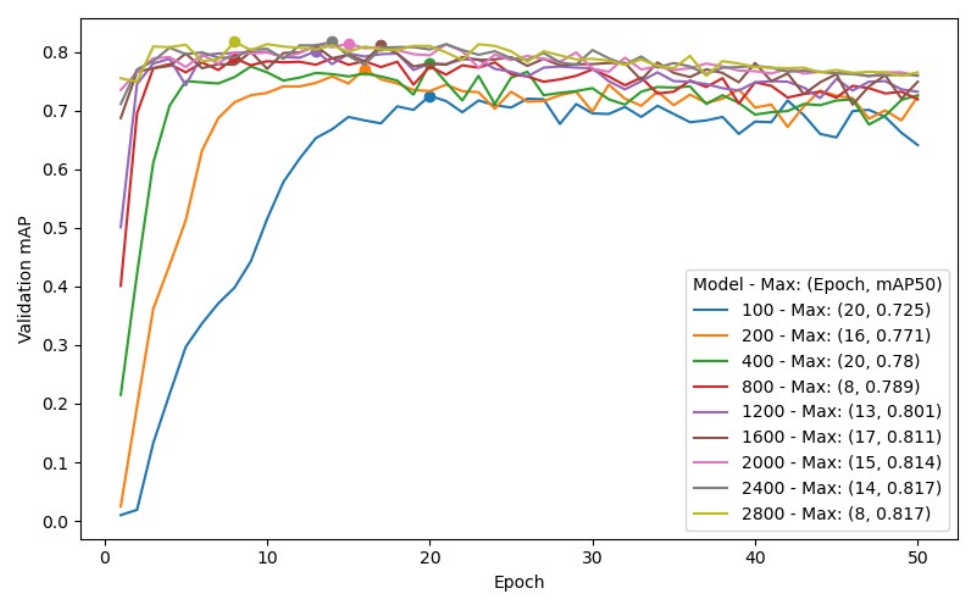}
\caption{Validation  accuracy of the GeoAI model across different training epochs with varying sample sizes. ``Max: (Epoch, mAP50)'' in the legend indicates the epoch at which the model achieves the highest predictive accuracy, measured by the standard metric mAP50 (mAP: mean average precision, 50 means a threshold setting in the measure). The corresponding maximum values are also highlighted on the performance curves as dots.} \label{fig_exp1valacc}
\end{figure}

\begin{figure}
\centering
\includegraphics[width=\textwidth]{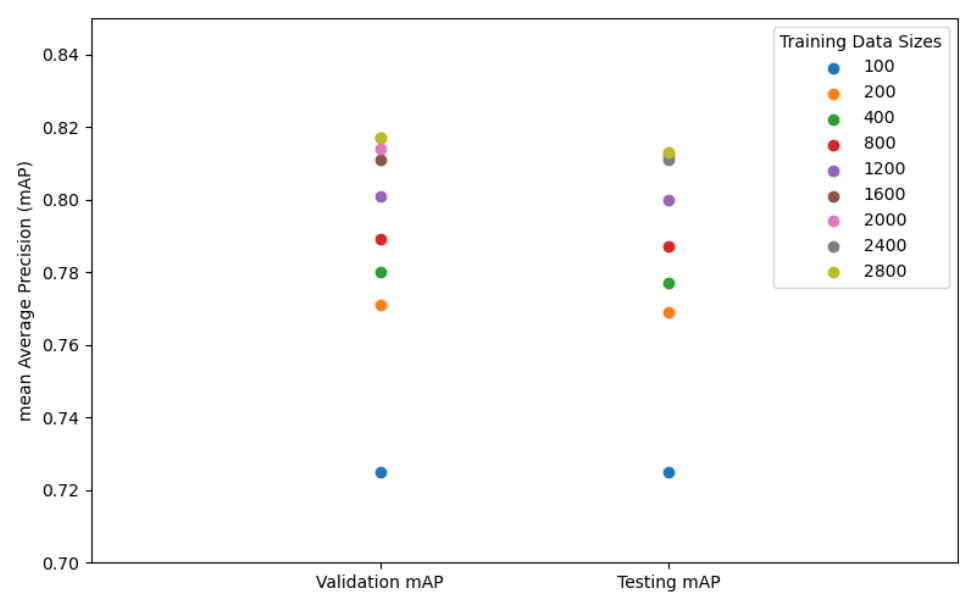}
\caption{The highest validation and testing accuracy (measured by mAP50) of GeoAI models trained with different training dataset sizes.} \label{fig_exp1testacc}
\end{figure}

The results depicted in Figure \ref{fig_exp1valacc} highlight the impact of training data volume on the rate of model accuracy improvement over time. A substantial increase in the prediction accuracy is observed for models trained with larger datasets, especially at the early stages of training, indicating that access to a more diverse set of images accelerates the learning process. However, a performance plateau is reached by all models, suggesting a limit to accuracy gains solely through increased epoch count, pointing to other potential limiting factors such as the model architecture or overfitting. Figure \ref{fig_exp1testacc}'s scatter plot complements these insights by comparing the predictive accuracy on both validation and testing datasets across the models. While accuracy improves with larger datasets, the incremental gains taper off, implying a saturation point in data utility. 

From the perspective of computational reproducibility, a discrepancy in the results becomes evident when the model is trained with different data sizes. For instance, as shown in Figure \ref{fig_exp1valacc}, there is an almost 10\% difference in prediction accuracy (mAP50) when the same model is trained with 2,800 images (mAP50: 0.817) compared to 100 images (mAP50: 0.725). The speed at which the model achieves its maximum performance also varies. Cross-comparing the model’s predictive accuracy on training and testing datasets (Figure \ref{fig_exp1testacc}), we observe a general trend that the testing accuracy is slightly lower than the validation accuracy on this Mars crater dataset. These observations indicate that without a clear description of the data used in GeoAI research, including its size, distribution, and partition, achieving satisfactory reproducibility is challenging.

Considering the combined evidence from Figures \ref{fig_exp1valacc} and \ref{fig_exp1testacc}, the model trained with 2,000 images emerges as the most practical option. It achieves near-peak accuracy without the diminishing returns associated with larger datasets and avoids the unnecessary computational demands they entail. To further consolidate this analysis, we conducted paired-samples T-tests \citep{ross2017paired} to verify if the mean change in model validation accuracy, when using different sample sizes, is significantly different from zero. Table \ref{tbl_trainingsize} lists the p-values for the T-test results. The model with each sample size listed in Table \ref{tbl_trainingsize} was run 10 times to generate the statistics. It can be observed that the results are statistically significant for all the sample size pairs increased from (100, 200) to (1,600, 2,000), with all p-values being smaller than the 0.05 threshold. However, as the sample size continues to increase beyond 2,000, we can no longer observe statistically significant T-test results. This indicates that for models running with more than 2,000 samples, there is no obvious difference in the model's predictive performance. This model configuration (with 2,000 samples), therefore, results in an optimal balance between high accuracy and computational efficiency, making it the preferred choice for deployment where both factors are critical. All the subsequent R\&R experiments were conducted using 2,000 images to compose the training dataset. This data choice and the reasoning behind it also become important factors ensuring understanding of the research aim and design principles to ensure reproduction.

\begin{table}[hbt!]
\tbl{Paired-samples T test for model validation accuracy with different sample sizes. The numbers in parentheses show the sample sizes used to train two models, the results of which are then used to conduct the T-test.}
{\begin{tabular}{lcccc} 
\toprule
Sample size pair & (100,200) & (200,400) & (400,800) & (800,1200) \\ 

\midrule
P-value  & \(4.6\times 10^{-16}\) & \(2.86\times 10^{-7}\) & \(1.38\times 10^{-7}\) & \(3.97\times 10^{-9}\) \\

\bottomrule
\toprule
Sample size pair & (1200,1600) & (1600,2000) & (2000,2400) & (2400,2800) \\
\midrule
P-value & \(1.71\times 10^{-7}\) & 0.016 & 0.31 & 0.44 \\

\bottomrule
\end{tabular}}
\label{tbl_trainingsize}
\end{table}

\subsection{Random seed effects on model consistency}
A fundamental aspect of ensuring reproducibility and replicability in GeoAI, particularly in deep learning models, is understanding the role of random seed settings. Random seeds play a crucial role in various operations within deep learning applications, including the initialization of model weights, the shuffling of data, or the application of regularization techniques such as dropout \citep{baldi2014dropout}. These elements are inherently stochastic and can affect the training process and the resulting model performance. Consistent random seed settings ensure predictability and uniformity in these operations, leading to more reproducible results, while varying random seeds can introduce a degree of variability, potentially affecting the reliability and replicability of the model outcomes. This experiment aims to investigate how the fixing or varying random seeds impacts the performance of models trained for Martian crater detection. By analyzing the influence of random seeds, we aim to gain insights into the variability of model outcomes and establish methodologies for more reliable and repeatable GeoAI research.

\begin{figure}
\centering
\includegraphics[width=0.6\textwidth]{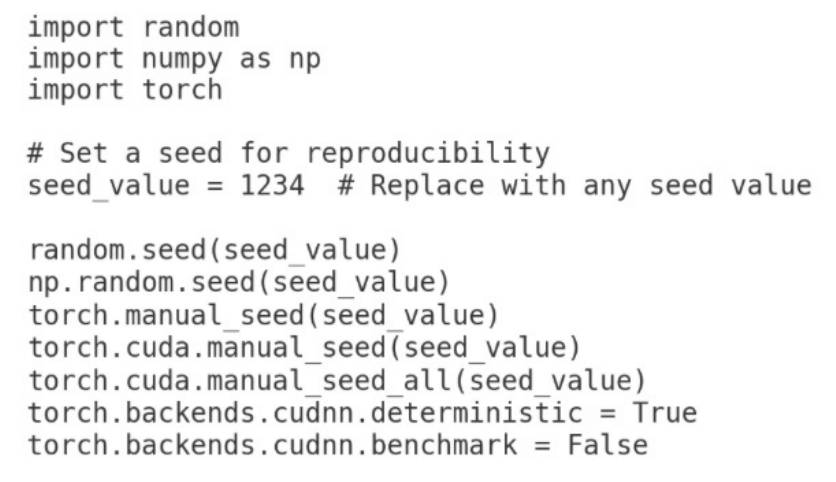}
\caption{Initialization of random seeds in Python to measure model reproducibility.} \label{fig_exp2code}
\end{figure}

Building upon the findings from Experiment 1, this experiment was conducted using a subset of 2,000 images from the Mars Crater Dataset for training, a size determined to be optimal based on our previous experimental results. Alongside this, 10,000 images were used for validation and another 10,000 for testing to ensure a comprehensive assessment of model performance. The first group in our experiment consisted of 20 runs of the GeoAI model with a fixed random seed, thereby ensuring identical initialization and data shuffling processes across training sessions. The second group included another 20 runs of models with unfixed random seeds, introducing variability in these initial conditions. The code used for fixing the random seed in our models is detailed in Figure \ref{fig_exp2code}. After the experiments, the mAP on both validation and testing sets for each model was calculated to compare performance variance caused by random seed settings.

\begin{figure}
\centering
\includegraphics[width=\textwidth]{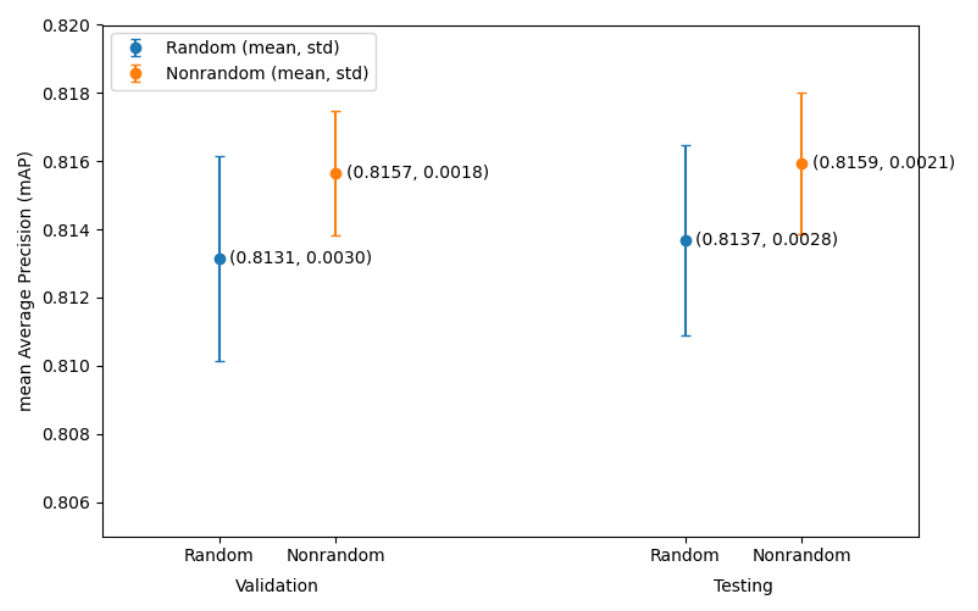}
\caption{Variance of the model performance with fixed (nonrandom) and random seed settings for both validation and testing datasets. \textit{mean} and \textit{std} refer to the average and standard deviation of prediction accuracy values over 20 model runs.} \label{fig_exp2result}
\end{figure}

The results of Experiment 2, as presented in Figure \ref{fig_exp2result}, provide a critical examination of the role random seed settings play in the stability of deep learning models with GeoAI tasks. To evaluate the consistency of model performance, Levene’s test \citep{leveneRobustTestsEquality1960} was employed – chosen for its robustness in assessing the equality of variances across different samples, which is less sensitive to departures from a normal distribution. The test revealed a statistically significant difference in variance, with p-values of 0.031 for validation and 0.041 for testing, both highlighting the importance of the random seed choice. These figures, below the alpha threshold of 0.05, suggest the models with a fixed random seed show more consistent performance, highlighting a key factor in achieving reproducible and replicable results. It is important to note, however, that fixing the random seed does not inherently improve performance (although in our experiments, it does yield better results); rather, it contributes to stability. The decreased standard deviation for the nonrandom group (fixed seed) seen in the figure points to a more predictable and dependable modeling process. By establishing that fixed random seeds contribute to more stable results, this experiment provides evidence that such practices are integral to the reproducibility and replicability of GeoAI models, confirming the ability to consistently reproduce findings across successive experiments.

\subsection{Locational variance on GeoAI model performance based on random sampling}
In the exploration of Martian landscapes, the ability of GeoAI models to produce consistent analytical results across varying spatial locations – a concept known as spatial replicability - is critical. This consistency fosters confidence in predictive models used across diverse geographic terrains and ensures that the insights derived are not merely byproducts of localized spatial anomalies. For instance, Martian craters exhibit a wide range of characteristics influenced by their geographical context; craters in the polar ice caps are characterized by layers of ice and dust deposits, contrasting with those in the equatorial region, which may display more pronounced erosion patterns due to wind activity \citep{edgett2003mars, kreslavsky2003north}. The spatial distribution of these features is intrinsically irregular and heterogeneous, further complicated by the varied angles and resolutions of data collection. Consequently, for GeoAI applications in planetary science and navigation requiring high precision---such as landing site selection or geological mapping---replicating findings across these diverse Martian terrains is essential. Experiment 3 is designed to evaluate this aspect of GeoAI models by analyzing their performance variance across a systematically segmented grid on the Martian surface.

\begin{figure}
\centering
\includegraphics[width=\textwidth]{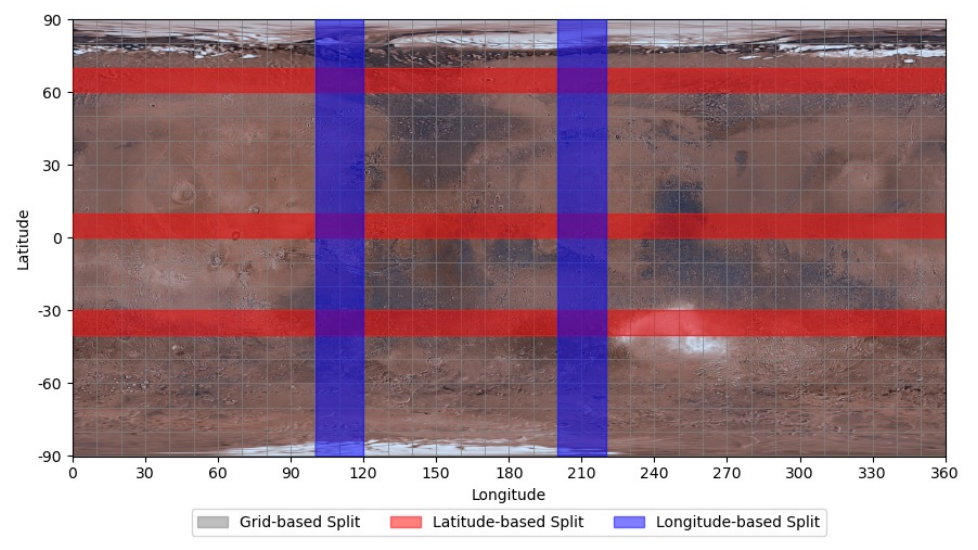}
\caption{Overview of grid-, latitude-, and longitude-based splits for R\&R analysis in GeoAI.} \label{fig_exp3split}
\end{figure}

In this experiment, we partitioned Mars surface into 10-degree by 10-degree grid cells (marked in gray lines in Figure \ref{fig_exp3split}). The objective is to assess the consistency with which our GeoAI model predicts crater characteristics across these diverse spatial segments. Drawing on the conclusions of previous experiments, Experiment 3 adopts a methodology that leverages the optimal sample size and the proven importance of random seed settings. The model was trained using a select sample of 2,000 images from across Mars, a size determined from Experiment 1 to be conducive to stable and high-performing results.

\begin{figure}
\centering
\subfloat[Performance variance of GeoAI model by grid.\label{fig_exp3map}]{%
\resizebox*{\textwidth}{!}{\includegraphics{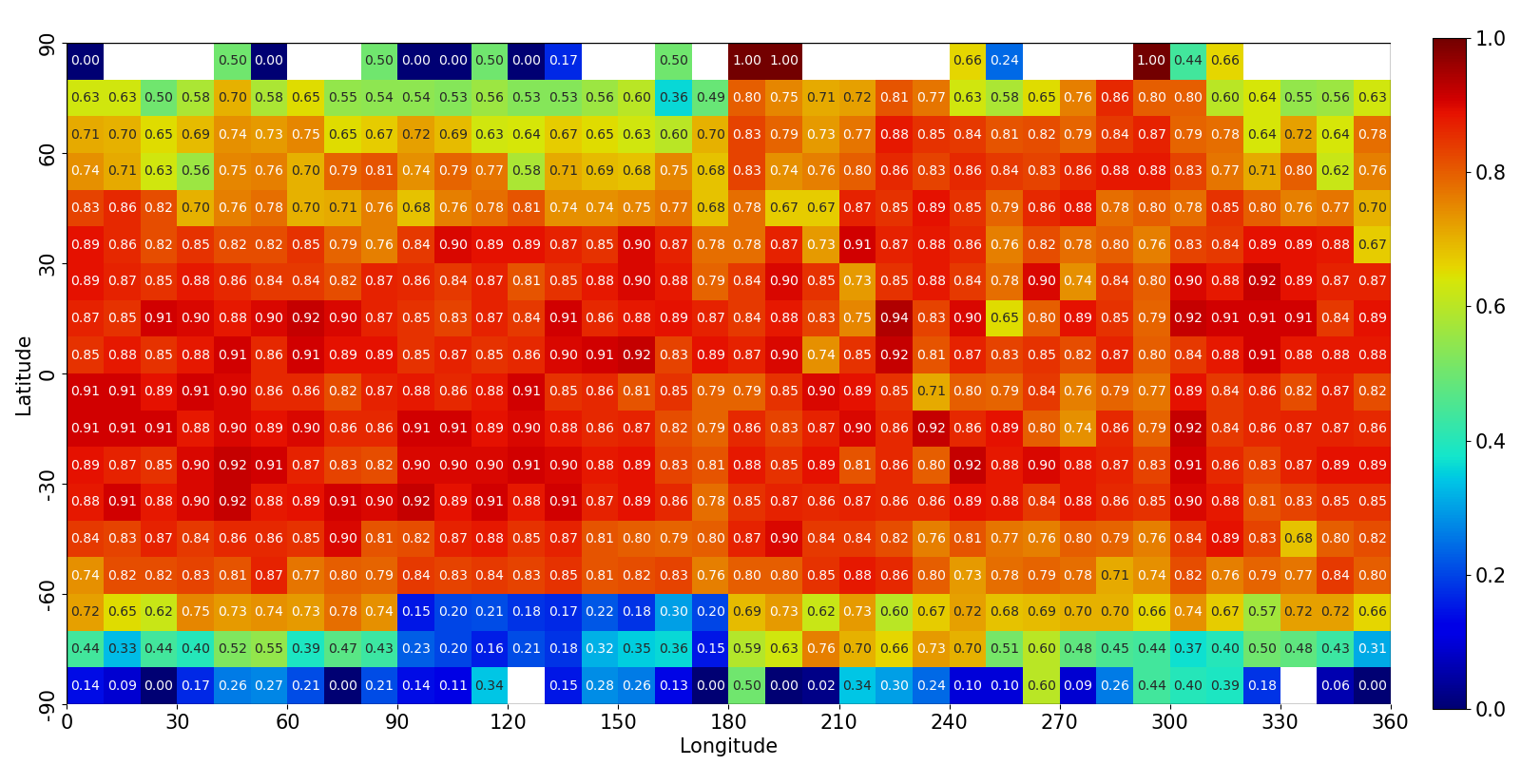}}}\hspace{5pt}
\subfloat[Amount of testing dataset per grid.\label{fig_exp3num}]{%
\resizebox*{\textwidth}{!}{\includegraphics{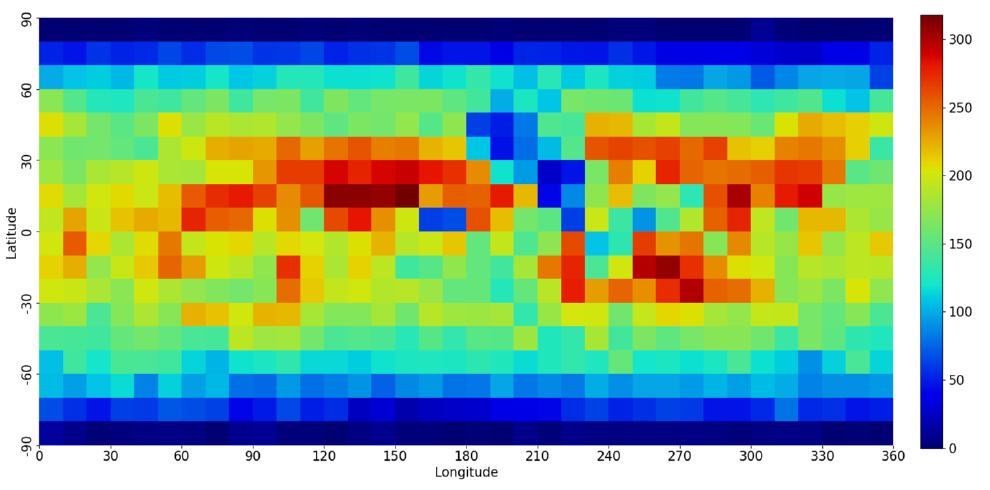}}}
\caption{Variance of GeoAI model performance (measured by mAP) (a) and amount of testing dataset (b) per grid cell.} \label{fig_exp3result}
\end{figure}

Figure \ref{fig_exp3map} presents the mAP scores for Martian crater detection across different grid cells, with scores ranging from 0 (indicating incorrect predictions for all images within the grid cell) to 1 (representing a perfect prediction). White grids denote areas with no data, possibly due to the absence of images or craters. From this result, or what we can call a form of a ``replicability map,'' it becomes apparent that even when a model is trained with images sampled globally, its spatial reproducibility across different geographical regions (the grid cells herein) still exhibits significant variance. Overall, regions near the equator demonstrate a higher degree of result reproducibility, as indicated by their elevated prediction accuracy values. In contrast, regions near the polar areas exhibit a lower degree of reproducibility, with a few exceptions where there is only one image scene in a grid (near the north pole), and the model makes the prediction correctly, resulting in a prediction accuracy of 1. This result reflects the pervasive spatial heterogeneity across the Martian surface, posing a challenge to a model’s spatial reproducibility.

Figure \ref{fig_exp3num} illustrates the statistics of the number of testing images distributed across the grid cells. This distribution closely aligns with that of the total images within each grid cell, as the testing set comprises all available images within each grid, excluding those designated for training or validation, which are fixed in size. Additionally, a noticeable pattern emerges: there are more images near the equator than the poles. This discrepancy arises because each grid cell covers a larger area near the equator than those near the poles, resulting in a greater number of ``valid'' image scenes available for training and testing. Through a Pearson correlation analysis, we found a high correlation coefficient of 0.71 between cell values in Figure \ref{fig_exp3map} and Figure \ref{fig_exp3num}, with a very low p-value (almost 0), indicating the statistical significance of this result. This outcome likely reflects the increased likelihood of data from image-rich regions being sampled and included in the training data, making the model more representative of those regions and resulting in higher prediction accuracy in the same areas. This analysis emphasizes the significance of data richness and representativeness, which is largely driven by spatial heterogeneity, in GeoAI research. It also reiterates the need for providing detailed descriptions of sampling strategies and data distribution about the original, training and testing data to enable the audience to gain a more comprehensive understanding of the likelihood of reproducibility for a GeoAI model. 

\subsection{Locational variance in GeoAI model performance across varying latitude}
Building upon our previous grid-based analysis, Experiment 4 introduces a latitude-based training/testing region partition approach to further investigate the locational impact of GeoAI model performance across varying latitudes on the Mars surface. In this experiment, we segment the Martian surface based on latitudes, with each strip being 10 degrees in size. Hence, the top strip (shown in Figure \ref{fig_exp3split}) represents the region between the 90th parallel north and the 80th parallel north, and the strip right below the equator is the region between the equator and the 10th parallel south. We randomly sampled 2,000 training images in three strips (highlighted in red in Figure \ref{fig_exp3split}) respectively and trained three models, testing them on images from the rest of the latitudinal strips. This methodology allows us to explore how well the model generalizes across different latitudinal zones, each potentially possessing unique geographical and geological characteristics due to varying solar exposure, atmospheric conditions, and surface compositions. Unlike the grid-based approach, which provided a broad overview of spatial reproducibility, this latitudinal division aims to uncover the GeoAI model’s spatial replicability linked to specific environmental and topographical variances present at different latitudes on Mars.

\begin{figure}
\centering
\includegraphics[width=\textwidth]{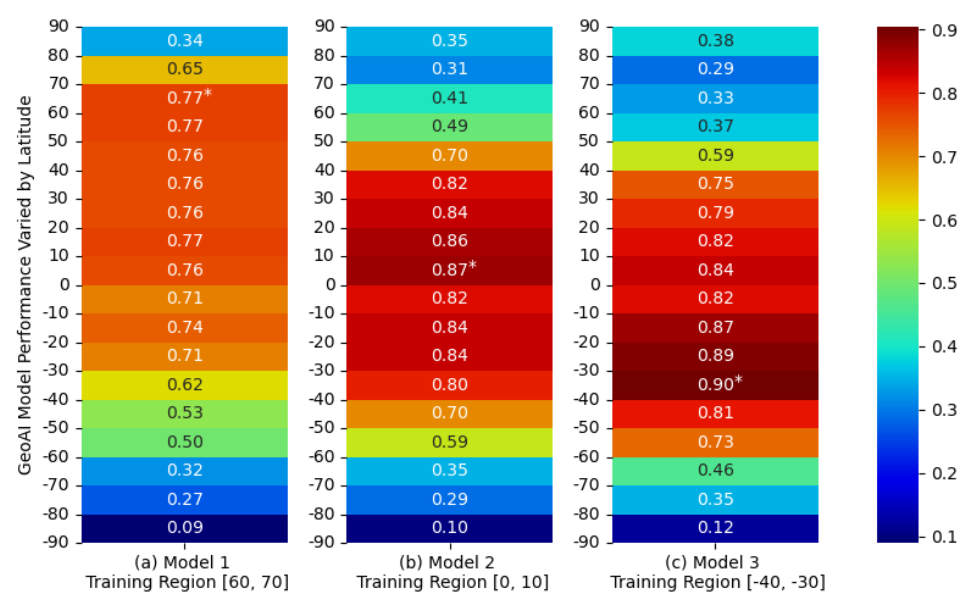}
\caption{GeoAI  performance variance in models trained from images in three latitudinal strips, which were (a) [60, 70], (b) [0, 10], and (c) [-40, -30] and tested in the others. * means images in this latitudinal strip are used for model training. The value next to * indicates the validation accuracy based on the model trained using images within the same region.} \label{fig_exp4result}
\end{figure}

Figure \ref{fig_exp4result} presents the results of our latitude-based analysis. The results are grouped into three bars, with each bar corresponding to the GeoAI model’s prediction accuracy when trained on images from the latitude strips [60, 70], [0, 10], and [-40, -30], and tested on the others. One can observe that when the model is trained and tested in the same region (indicated by a number followed by a star), it yields the highest prediction accuracy compared to when tested on images from other latitude strips. However, the model’s performance in the other testing regions cannot fully replicate this value. This is likely due to spatial heterogeneity, such as varying solar exposure and atmospheric conditions in latitude-varying regions. This experiment demonstrates that location factors, especially latitude, play a critical role in the spatial replicability of GeoAI research.

In the meantime, another pattern is observable, where the GeoAI model’s prediction accuracy is higher and closer to the highest score in each bar when the test region is in closer proximity to the training region. According to Tobler’s first law of geography, which states that closer things are more related to each other, we can infer that the data characteristics are similar for regions that are closer in latitude; therefore, the model exhibits higher spatial replicability. However, the degree of spatial replicability is not only related to spatial dependence in the data but also to the number of testing datasets, as explained in Section 4.3. For the first bar in Figure \ref{fig_exp4result}, even though the top strip between latitude parallels [80, 90] is only one strip further from the training region located between latitude parallels [60, 70], due to the sparsity of data in that region, the spatial replicability (reflected by predictive accuracy) is lower than those strips further away from the training region but down in latitudes, such as the testing region within latitudes [0, 10].

\begin{table}
\tbl{Moran’s I statistics on the spatial variance of the GeoAI model’s prediction accuracy. Models 1, 2, and 3 are trained on images collected between latitude parallels [60, 70], [0, 10], and [-40, -30] respectively.}
{\begin{tabular}{l >{\centering\arraybackslash}p{0.1\textwidth} >{\centering\arraybackslash}p{0.1\textwidth} >{\centering\arraybackslash}p{0.1\textwidth}} 
\toprule
\multirow{3}{*}{Spatial statistics} & Model 1 & Model 2 & Model 3 \\
\cmidrule{2-4}
& \multicolumn{3}{l}{Locations (latitude range) for training strips} \\
\cmidrule{2-4}
& [60, 70] & [0, 10] & [-40, -30] \\
\midrule
Moran’s I & 0.76 & 0.87 & 0.85 \\
P-value & 0.001 & 0.001 & 0.001 \\
Z-score (simulated) & 3.72 & 3.85 & 3.82 \\

\bottomrule
\end{tabular}}
\label{tbl_exp4}
\end{table}

To confirm our findings, the Moran's I statistic is applied to all testing results from the three bars shown in Figure \ref{fig_exp4result}. Each bar refers to the model’s predictive accuracy across space based on models trained on images from a certain latitude strip. As the results in Table \ref{tbl_exp4} show, the values of Moran’s I analyzed on the three models’ output are all positive and close to 1, indicating a strong positive spatial autocorrelation. The prediction results for Model 2, which was trained on a latitude strip near the equator, exhibit the highest level of spatial autocorrelation compared to the other two. The associated p-values, all marked at 0.001, indicate a very low probability of such a spatial pattern occurring by chance, reinforcing the significance of the observed spatial clustering. The simulated Z-scores, exceeding the critical value of 1.96, confirm the robustness of the results, with scores of 3.72, 3.85, and 3.82 respectively, underscoring the substantial deviation from a null hypothesis of random distribution. Collectively, these statistical measures validate the visual observation of the spatial dependency among the replication results across latitudes, with a tendency to have a full replication in areas that are closer to where the training dataset comes from.

\subsection{Locational variance on GeoAI model performance across varying longitude}

Complementing our latitude-based partition, Experiment 5 further advances our understanding of GeoAI’s spatial replicability by undertaking a longitude-based training/testing region partition. The shift to a longitude-based partition is driven by the need to investigate the influence of distinct geophysical and environmental variations along Martian meridians. Longitude-induced variations are characterized by differences in topography, mineral composition, and the imprints of historical geological activities, setting them apart from the primarily climatic and polar influences observed in latitudinal changes. Such variations can be decisive in data representation and, consequently, the performance of GeoAI models tasked with crater detection. By adapting our analysis to the longitude-based approach, we aim to discern whether and how the same replicability issues can be observed across regions varying by longitude.

In this experiment, we adapt the data partitioning strategy from latitude-based to longitude-driven. As shown in Figure \ref{fig_exp3split}, we segment the Martian surface every 20 degrees by longitude lines (meridians) to create distinct training and testing zones. The choice of a 20-degree space, as opposed to a 10-degree space, for partitioning the Mars surface is to accumulate a sufficient number of images for both training and validation. The blue strips in Figure \ref{fig_exp3split} denote the two regions (between 100-120 degrees and 200-220 degrees) chosen to sample images to train two GeoAI models. The rest of the 20-degree wide strips are used for testing. These two regions were selected based on image availability, with the former containing the most images at 6,293 and the latter having the least at 4,388. By training models on these two diverse regions---one image-rich and one image-poor---we aim to examine the model's performance against extremes of data density and understand how data variances across longitude would affect GeoAI detection capability.

\begin{figure}
\centering
\includegraphics[width=\textwidth]{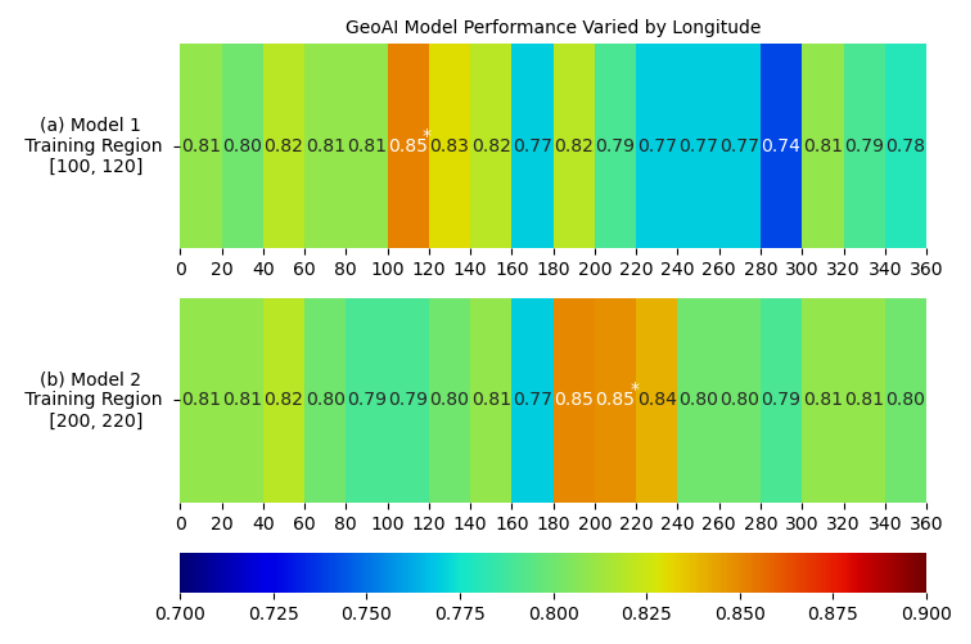}
\caption{GeoAI  performance variance in models trained from images in two strips partitioned by meridians, which are between (a) [100, 120] and (b) [200, 220] meridians and tested in the others. The values reflect the model’s prediction accuracy based on mAP50. * means images in this longitude strip are used for model training. The value next to * indicates the validation accuracy based on the model trained using images within the same region.} \label{fig_exp5result}
\end{figure}

\begin{table}
\tbl{Moran’s I   statistics on the spatial variance of the GeoAI model’s prediction accuracy. Models 1 and 2 are trained on images collected between [100, 120] and [200, 220] in longitude, respectively.}
{\begin{tabular}{l >{\centering\arraybackslash}p{0.18\textwidth} >{\centering\arraybackslash}p{0.18\textwidth}} 
\toprule
\multirow{3}{*}{Spatial statistics} & Model 1 & Model 2 \\
\cmidrule{2-3}
& \multicolumn{2}{l}{Locations (latitude range) for training strips} \\
\cmidrule{2-3}
& [100, 120] & [200, 220] \\
\midrule
Moran’s I & 0.39 & 0.23 \\
P-value & 0.022 & 0.128 \\
Z-score (simulated) & 1.93 & 1.20 \\

\bottomrule
\end{tabular}}
\label{tbl_exp5}
\end{table}

Figure \ref{fig_exp5result} presents the results of the replication analysis for GeoAI models with training/testing region partition based on longitude. The two training strips are regions between [100, 120] and [200, 220] meridians, respectively. Unlike the replication results from the latitude-based analysis, a visual inspection of the figure does not immediately suggest a spatial dependency of accuracy scores based on the proximity to the training strip. The distribution of prediction accuracy scores across longitude strips appears more uniform, lacking the marked gradient observed in the latitude-based analysis. For instance, the model's prediction score for the longitude strip between [160,180] is lower than those for other strips. Our examination has identified a combination of factors that could be contributing to these results, including the complexity of the local terrain, crater morphology, and the quality of the image mosaic.

The spatial autocorrelation statistics for the model’s predictions across space (longitude) based on the two models trained with images from different longitude strips are shown in Table \ref{tbl_exp5}. The two Moran’s I scores, 0.39 and 0.23, are notably lower than those observed in the latitude-based analysis, where Moran's I values were closer to 1, indicating only weak positive spatial autocorrelation. The p-values of 0.022 and 0.128 for the spatially induced performance variance analysis of the two models, along with simulated Z-scores of 1.93 and 1.20, further indicate a weaker and statistically less significant spatial autocorrelation in this longitude-based analysis. The p-value for the result of Model 2 exceeds the conventional threshold of 0.05, and the Z-score does not meet the standard criterion for significance, suggesting no significant spatial pattern.

These findings imply that, for this particular set of models and data, the variations in the GeoAI model’s performance caused by longitude do not exhibit strong spatial autocorrelation patterns, unlike the patterns observed in the latitude-based analysis. Instead, spatial heterogeneity plays a more central role, as we can hardly observe a full spatial replication for both models in Figure \ref{fig_exp5result}, except for Model 2 when tested on the longitude strip [180, 200]. Thus, the model’s spatial replicability across longitude shows a more random pattern.

\section{Conclusion}
This paper provides a comprehensive view of the factors influencing the computational and spatial replicability of GeoAI research. Computationally, various sources of model uncertainty, such as random seed settings, and the implicit description of the use of training data and methods, hinder the full reproducibility of results in GeoAI. Due to the complexity of model architecture, its handling of geospatial big data, and its intricate training, fine-tuning, and deployment processes, GeoAI research is more challenging to reproduce than other shallow machine learning or statistical approaches. In addition to summarizing the computational and modeling factors that hinder result reproducibility, we discussed the importance of open science and detailed technical documentation in improving the computational reproducibility of GeoAI research. 

In addition to computational challenges, we present a unique perspective on the crucial role of location in the spatial replicability of GeoAI methods. This reflects a higher level of scientific challenge in reproducibility and replicability (R\&R) research. Using crater detection on Mars as an example, we conducted analyses to partition training and testing regions based on grids, latitude, and longitude. Through both visual inspection and statistical examination of the ``reproducibility maps,'' our study reveals that the spatial replicability of GeoAI research, in the case of crater detection, is not uniform and varies significantly by location. In fact, the uncontrolled variance of Earth's surface indicates that, despite controlling all variables and adhering to a well-documented research workflow (including data, methods, system environment, software, etc.), achieving complete replicability may still be unattainable. This observation supports the concept of `weak replicability' emphasized by \citet{goodchildReplicationSpaceTime2021} in environmental and social research using spatial methods.

An intriguing finding from our latitude and longitude-based analyses is that spatial autocorrelation and spatial heterogeneity play pivotal roles in the degree of spatial replicability in GeoAI research. In the Mars crater detection case, we observed smaller result variances when a testing region is in closer proximity to the original study area in the latitude-based analysis. The results also demonstrate a strong spatial clustering pattern based on Moran’s I analysis, indicating a higher similarity in data characteristics along latitudes. However, the longitude-based analysis did not reveal the same spatial autocorrelation patterns in the reproduced results. Instead, spatial heterogeneity, such as topography, mineral composition, and the imprints of historical geological activities, plays a more central role in determining longitude-based spatial replication. 

Based on this research, several directions are worth further investigation in GeoAI research, particularly regarding its reproducibility and replicability. First, given that ``weak replicability'' is an inherent property of spatial methods, including GeoAI, it is crucial to discuss the spatial replicability of proposed methods in other study areas and its implications. Before drawing definitive conclusions, testing models across geographically heterogeneous regions and locations is essential. Simultaneously, the development of measures for ``out-of-distribution'' is critical to quantify the similarity in distribution patterns and data characteristics between training and test datasets, providing a better understanding of replicability across space and time. 

Second, map projections and grid division strategies are important considerations when conducting geospatial research, especially at a global scale. As the Mars image mosaic adopts a cylindrical projection, it causes significant distortions to the craters near the polar regions. To address this challenge, we ensured that in our data partitioning, each image scene covers the same area, and distortion was corrected algorithmically before an image scene was used for training \citep{hsuKnowledgeDrivenGeoAIIntegrating2021}. On a related matter, grid partitioning also affects the area covered by each grid. In our case, as presented in Section 4.3, the partition by latitude and longitude causes the grid area to cover a smaller area near the polar regions than at the equator. Our experiments specifically analyzed the impact of this, and the correspondingly different number of craters within each grid, on the reproducibility of the GeoAI modeling results. In Section 4.4, the strip partition along latitude encounters the same issue, and we developed strategies to mitigate this area effect by selecting the same number of training images per strip for the analysis. Hence, the impact of location along latitude can be more clearly demonstrated. To further understand the impact of grid partitioning strategies, one interesting future research topic is to conduct hexagon-based or triangular-based partitioning, ensuring each hexagon or triangle covers the same area. In this case, we can better observe the crater distribution across the Martian surface and understand the location transfer effect (experiments in Section 4.3) and how it differs from using a latitude-longitude-based partition method. The dynamic visualization of such replicability variance will enrich the implementation strategies for the ``replicability map,'' which is critical to improve our understanding of GeoAI R\&R from cognitive perspectives. 

Third, fostering a culture of sharing should go beyond code and data to include design principles, the underlying choices made for both data and GeoAI models, as well as the trial-and-error aspects of the research process. Ensuring a deep understanding of the research process beyond what a paper reports is critical for the meaningful adaptation of models and further innovations. This value of sharing research experience and lessons learned extends beyond simple reproduction.

\bibliographystyle{apacite}
\bibliography{interactapasample}

\end{document}